\documentclass[a4paper]{article}
\usepackage{iwslt18}
\ninept

\usepackage{graphicx}
\graphicspath{ {figs/} }

\usepackage{amssymb}
\usepackage{amsmath}
\usepackage{epsfig}

\usepackage{dsfont}
\usepackage{mathtools}
\usepackage{xcolor}
\usepackage{multirow}

\usepackage[marginpar]{todo}

\usepackage{tikz}
\usetikzlibrary{calc,arrows.meta}

\tikzset{embedding/.pic={
\fill[color=black] (-6pt,-11pt) rectangle (6pt,-1pt);
\fill[color=white] (-3pt,-6pt) circle[radius=2pt]
                   (3pt,-6pt) circle[radius=2pt];}}

\tikzset{hidden/.pic={
\fill[color=black,rounded corners] (-6pt,-11pt) rectangle (6pt,-1pt);
\fill[color=white]  (-3pt,-6pt) circle[radius=2pt]
                    (3pt,-6pt) circle[radius=2pt];}}

\tikzset{softmax/.pic={
\fill[color=black,rounded corners] (-6pt,-11pt) rectangle (6pt,-1pt);
\fill[color=white] (0pt,-6pt) circle[radius=2pt];}}

\newcommand{\model}[1]{\texttt{#1}}

\begin{document}
\title{Neural Baselines for Word Alignment}

% Submission is anonymous
%\name{Anonymized submission }
%\address{}

\makeatletter
\def\name#1{\gdef\@name{#1\\}}
\makeatother
\name{{\em Anh Khoa Ngo Ho, Fran\c{c}ois Yvon}}
\address{LIMSI, CNRS, Universit\'e Paris-Saclay \\ B\^{a}t. 508, rue John von Neumann, Campus Universitaire, F-91405 Orsay \\
{\small \tt anh-khoa.ngo-ho@limsi.fr, francois.yvon@limsi.fr}
}

\maketitle

\thispagestyle{empty}

\begin{abstract}
Word alignments identify translational correspondences between words in a parallel sentence pair and is used, for instance, to learn bilingual dictionaries, to train statistical machine translation systems, or to perform quality estimation. In most areas of natural language processing, neural network models nowadays constitute the preferred approach, a situation that might also apply to word alignment models.
In this work, we study and comprehensively evaluate neural models for unsupervised word alignment for four language pairs, contrasting several variants of neural models. We show that in most settings, neural versions of the IBM-1 and hidden Markov models vastly outperform their discrete counterparts. We also analyze typical alignment errors of the baselines that our models overcome to illustrate the benefits --- and the limitations --- of these new models for morphologically rich languages.
\end{abstract}

\section{Introduction\label{sec:introduction}}
Word alignment is one of basic tasks in multilingual Natural Language Processing (NLP) and is used to learn bilingual dictionaries, to train statistical machine translation (SMT) systems, to filter out noise from translation memories or in quality estimation applications \cite{Specia18quality}. Given a pair of sentences consisting of a sentence in a source language and its translation in a target language, word alignments aims to identify translational equivalences at the level of individual word tokens \cite{Och2003A-Systematic,Tiedemann2011Bitext}. Until recently, the most successful alignment models were statistical, as represented by the IBM Models \cite{Brown1993Mathematics} and the HMM model \cite{Vogel1996HMM-based}. These models use unsupervised estimation techniques to build alignment links at the word level, relying on large collections of parallel sentences.

Such approaches are typically challenged by low-frequency words, whose cooccurrences are poorly estimated; they also fail to take into account context information in alignment; finally, they make assumptions that are overly simplistic (eg.\ that all alignments are one-to-many or many-to-one), especially when the languages under focus belong to different linguistic families. Even though their performance seems fair for related languages (eg.\ French-English), there is still much room for improving automatic alignments produced by standard tools such as \model{Giza++} \cite{Och2003A-Systematic} or \model{Fastalign} \cite{Dyer2013A-Simple}.

% \fy{outdated ?}
As is the case for most NLP applications \cite{Collobert2011Natural}, and notably for machine translation (MT) \cite{Cho2014On-the-Properties, Bahdanau2014Neural, Luong15effective}, neural-based approaches offer new ways to address some of these issues. One important reason for this success is the implicit feature extraction performed by neural networks, which represents each word as a dense low-dimensional vector and effectively extends word representations by vector concatenation \cite{Young2017Recent}. Following up on the work of \cite{Yang2013Word,Tamura2014Recurrent,Alkhouli16alignmentbased,Wang17Hybrid,Wang2018Neural}, we focus here word alignments, trying to precisely assess the benefits of neuralizing standard word alignment models.

To this end, we design and implement multiple neural variants of the IBM and HMM models, and experiment with four language pairs (eight directions), and also consider multiple data conditions. In our analysis, we not only report improved AER scores, but also detail the positive impact of these neural baselines on standard alignment error types such as aligned and non-aligned words, rare vs.\ frequent words, etc. We also discuss the relevance of our neural network variants for each language pair and error type.
We therefore make the following contributions:
\begin{itemize}
\item a systematic comparison of several neural models for word alignments including context-independent models, contextual models and character-based models, which allow us to establish strong baselines for further studies.
\item a detailed error analysis for four European language pairs: English with French, German, Czech and Romanian.
\end{itemize}
Our experiments notably reveal that neuralized versions of standard alignment models vastly outperform their discrete counterparts, but also show that there still exists much room for improvements, especially when dealing with morphologically rich languages or in low-resource settings.

\section{Neural Word alignment models\label{sec:word_alignment_model}}

\subsection{Statistical Word alignment\label{ssec:word_alignment}}
The standard approach to statistical alignment \cite{Och2003A-Systematic} is to consider \emph{asymmetric} models associating each word in a source sentence $f_1^J = f_1\dots{} f_J$ of $J$ words with exactly one word from the target sentence $e_1^I = e_0 \dots{} e_I$ of $I+1$ words.\footnote{As is custom, the target sentences is completed with a ``null'' symbol, conventionally at index $0$. Our implementation is slightly more complex (see details in Section~\ref{ssec:transition_model}).} This relationship can be modeled as:
\begin{multline}
P(f_1^J|e_1^I) = \sum_{a_1^J} P(f_1^J, a_1^J|e_1^I) \\
= \sum_{a_1^J}  \prod_{j=1}^J p( f_j | f_1^{j-1}, a_1^{j}, e_1^I) \times{}  p( a_j | f_1^{j-1}, a_1^{j-1}, e_1^I)
\label{eq:posterior_prob}
\end{multline}
where $a_1^J = a_1 \dots{}a_J$ are the latent alignment variables, with $a_j \in [0 \ldots I]$. The two terms in the inner product in equation~\eqref{eq:posterior_prob} are referred to respectively as the \emph{translation model} and the \emph{alignment model}.

\subsection{Neural Translation Models\label{ssec:models}}

Both \model{IBM-1} and \model{HMM} make the simplifying assumption that $p(f_j | f_1^{j-1}, a_1^{j}, e_1^I)$ simplifies to $p(f_j | e_{a_j})$.  Analogous to these models, we propose two baseline neural variants \model{IBM-1+NN} and \model{HMM+NN}, where we implement the translation component with a neural network. As explained below, we then develop several additional versions, all relying on a simple and computationally efficient feed-forward architecture.

\subsubsection{A Baseline neural model\label{ssec:Translation_model}}

Our first neural model only modifies the translation model, keeping the transition model unchanged with respect to the corresponding baseline. Both the \model{IBM-1+NN} and \model{HMM+NN} use a simple feed-forward architecture which computes a distribution over possible source words $f_j$ from an input target word $e$. This is implemented as a single linear layer, followed by a softmax layer. In this architecture, a fixed size target vocabulary has to be specified to compute the softmax.
\begin{eqnarray}
    p_{\theta}(f_j | f_1^{j-1}, a_1^{j}, e_1^I) = p_{\theta}(f_j | e_{a_j})
\end{eqnarray}

In this framework, EM also applies \cite{Bergkirkpatrick10painless,Tran2016Proceedings}: during the (E) step, alignment posteriors are computed as usual using the Baum-Welsh algorithm; in the (M) step, the main change is that the NN parameters have to be optimized numerically, eg.\ via gradient descent.

\subsubsection{A Contextual translation model\label{ssec:Contextual_translation_model}}

A first variant adds some context around the target word.
% , which might prove useful when aligning morphologically complex words (in the source) to noninflected target English words.
As the target words are fully observed, this modification has no impact on the computations needed to implement the model. 
We use a sliding window of size $(2*h + 1)$ to represent word contexts and model $p(f_j | f_1^{j-1}, a_1^{j}, e_1^I)$ as $p(f_j |a_j, e_{a_j-h}^{a_j+h})$. For this variant, we compare two approaches to combine the embeddings of words in the context window:
\begin{itemize}
\item Concatenation (\model{NN+CtxCc}): we concatenate all word embeddings inside a window of size $h$ and use a feed-forward layer for combination. We consider that the context of the null ``word'' is made of null tokens, similarly to \cite{Yang2013Word}.
  % \fy{Why not use the context as well ?} 
\item Convolution (\model{NN+CtxCnn}): we use a convolution filter of size $(2*h +1, 2*h +1)$ to combine context words. We use a simpler approach for the null model by performing a convolution over a window of null tokens.
\end{itemize}

% \begin{figure}[!h]
% \centering
% \begin{tikzpicture}
% \node (E1) at (-60pt, 0pt) {$e_1$};
% \node (E2) at (-22pt, 0pt) {$e_2$};
% \node (E3) at (22pt, 0pt)  {$e_3$};
% \node (E4) at (60pt, 0pt)  {$e_4$};

% \draw[draw=black] (-30pt,-4pt) rectangle ++(100pt,10pt);
% \pic [local bounding box=N1] at (-60pt, -10pt) {embedding};
% \pic [local bounding box=N2] at (-22pt, -10pt) {embedding};
% \pic [local bounding box=N3] at (22pt, -10pt) {embedding};
% \pic [local bounding box=N4] at (60pt, -10pt) {embedding};
% \node[align=left] (LookupLayer) at (-98pt, -15pt) {\small Lookup \\ \small layer};

% \draw[-latex] (E1)  -- (N1);
% \draw[-latex] (E2) -- (N2);
% \draw[-latex] (E3) -- (N3);
% \draw[-latex] (E4)  -- (N4);

% \pic [local bounding box=L3] at (22pt, -25pt) {hidden};

% \draw[-latex] (N2) -- (L3);
% \draw[-latex] (N3) -- (L3);
% \draw[-latex] (N4) -- (L3);

% \pic [local bounding box=M3] at (22pt, -40pt) {hidden};
% \draw[-latex] (L3) -- (M3);
% \node[align=left] (HiddenLayer) at (-98pt, -40pt) {\small Hidden \\ \small layers};

% \pic [local bounding box=O3] at (22pt, -55pt) {softmax};
% \draw[-latex] (M3) -- (O3);
% \node[align=left] (HiddenLayer) at (-95pt, -60pt) {\small Softmax};

% \node (F3) at (22pt, -75pt) {$p(f|e_2, e_3, e_4)$};

% \end{tikzpicture}
% \caption{Structure of the neural contextual translation model}
% \label{fig:Contextual_translation_model}
% \end{figure}

\subsubsection{Character-based representations\label{ssec:Using_character-based_representations}}

We consider ways to use character-based representations to improve or even replace word embeddings, so as to accommodate arbitrary vocabularies in source and target. We apply a Bi-LSTM model to encode all characters in a target word $e$ respectively in the forward $\overrightarrow{h_{e}}$ and backward $\overleftarrow{h_{e}}$ direction. We concatenate the resulting two hidden states $[\overrightarrow{h_{e}}, \overleftarrow{h_{e}}]$ to represent each target word. Again, three variants are considered:
\begin{itemize}
\item Pure character-based representations on the target side \model{NNCharTgt} ; % \fy{The equations are worthless} % $p(f_j | f_1^{j-1}, a_1^{j}, e_1^I) = p(f_j | e_{a_j,1}^{a_j,l_{e_{a_j}}})$.
\item Combined character-based and word-based representations on the target side \model{NNCharWord}, where we simply concatenate both representations; % $p(f_j | f_1^{j-1}, a_1^{j}, e_1^I) = p(f_j | [e_{a_j,1}^{a_j,l_{e_{a_j}}}, e_{a_j}])$.
\item Pure character-based representation on both sides \model{NNCharBoth}. % : $p(f_j | f_1^{j-1}, a_1^{j}, e_1^I) = p(f_j | [e_{a_j,1}^{a_j,l_{e_{a_j}}}, f_{j,1}^{j,l_{f_{j}}}])$.
  While the first two variant only amount to changing the target embeddings, this latter model is more challenging as we modify the source embeddings that are used in output layer. While we keep a fixed size target vocabulary in the softmax computation during training, we are in a position to compute the association of any source with any target word, known or unknown, during testing.

 % In detail, the probability of source word $f_j$ is generated by $softmax(W_{f_j} h_{e_{a_j}})$ where $h_{e_{a_j}}$ is the hidden state of target word and $W_{f_j}$ is the matrix including the hidden states of all source words in vocabulary. The character-based representation on both sides helps to eliminate unknown word.
\end{itemize}

%\begin{figure}[!h]
%\centering
%\begin{tikzpicture}
%\node (E1) at (-60pt, 0pt) {c};
%\node (E2) at (-22pt, 0pt) {a};
%\node (E3) at (22pt, 0pt)  {t};
%\node (E4) at (60pt, 0pt)  {s};
%
%\pic [local bounding box=N1] at (-60pt, -10pt) {embedding};
%\pic [local bounding box=N2] at (-22pt, -10pt) {embedding};
%\pic [local bounding box=N3] at (22pt, -10pt) {embedding};
%\pic [local bounding box=N4] at (60pt, -10pt) {embedding};
%\node[align=left] (LookupLayer) at (-98pt, -15pt) {\small Lookup \\ \small layer};
%\draw[-latex] (E1)  -- (N1);
%\draw[-latex] (E2) -- (N2);
%\draw[-latex] (E3) -- (N3);
%\draw[-latex] (E4)  -- (N4);
%
%\pic [local bounding box=M1] at (-60pt, -25pt) {hidden};
%\pic [local bounding box=M2] at (-22pt, -25pt) {hidden};
%\pic [local bounding box=M3] at (22pt, -25pt) {hidden};
%\pic [local bounding box=M4] at (60pt, -25pt) {hidden};
%\node[align=left] (HiddenLayer) at (-95pt, -35pt) {\small Bi-LSTM};
%\draw[-latex] (N1)  -- (M1);
%\draw[-latex] (N2) -- (M2);
%\draw[-latex] (N3) -- (M3);
%\draw[-latex] (N4)  -- (M4);
%
%\draw[<->] (M1)  -- (M2);
%\draw[<->] (M2)  -- (M3);
%\draw[<->] (M3)  -- (M4);
%
%\pic [local bounding box=L3] at (0pt, -40pt) {hidden};
%\draw[-latex] (M1) -- (L3);
%\draw[-latex] (M4) -- (L3);
%
%\pic [local bounding box=M3] at (0pt, -55pt) {softmax};
%\draw[-latex] (L3) -- (M3);
%\node[align=left] (HiddenLayer) at (-95pt, -62pt) {\small Softmax};
%
%\node (F3) at (0pt, -75pt) {$p(f|cats)$};
%
%\end{tikzpicture}
%\caption{Structure of the character-based neural translation model}
%\label{fig:Character_translation_model}
%\end{figure}

\subsection{Alignment models\label{ssec:transition_model}}
\subsubsection{Baseline: a jump model}
We mostly follow the assumptions of \cite{Och2003A-Systematic} to design our alignment models. Only first-order dependencies are taken into account; furthermore, alignment  positions only depend on the jump width and not on the absolute index positions \footnote{We restrict ourselves to jump values in the interval $[-K,+K]$ where $K$ is a parameter of our model. For each sentence, the remaining probability mass corresponding to jumps greater than $K$ or lower than $-K$ is uniformly divided among those valid offsets \cite{Liang2006Alignment}. This means that we parameterize alignments using a multinomial distribution over $(2K + 3)$ buckets.}:
\begin{equation}
p( a_j | f_1^{j-1}, a_1^{j-1}, e_1^I) = p(\Delta_{a_{j}})\label{eq:jump} % = p(k)
\end{equation}
where $\Delta_{a_{j}} = a_j - a_{j-1}$.

Note that we associate a specific null token to every target word, which allows us to faithfully model jumps from and to null tokens. The probability to transition to an empty word is governed by one single parameter $p_0$. Constraints for transitioning into and out of empty words follow the proposal of \cite{Och2003A-Systematic}. For all variants of \model{IBM-1}, we thus use a uniform transition distribution $p(a_j | a_{j-1}) = \dfrac{1}{2I}$.

\subsubsection{Neural alignment models}
Our alignment models used in the \model{HMM} also rely on MLPs to compute the multinomial distribution in~\eqref{eq:jump}; they further combine character based representations for the word embeddings, as well as contextual word representations. Two variants are considered, where we only take the source, or the source and the target into account. 

\begin{itemize}
\item Character-based representation on the target side \model{NNJumpTgt}: here the jump value only depends on target words. We use the same character-based representations as above to represent words and also use a Bi-LSTM to encode target word contexts. Therefore, the alignment probability becomes:
\begin{equation}
p(a_j| f_{1}^{j-1}, a_1^{j-1}, e_1^I) = p(\Delta_{a_{j}} | h_{a_{j-1}}) % ,1}^{a_{j-1},l_{e_{a_{j-1}}}} )
\end{equation}
where $h_{a_{j-1}}$ combines the forward and backward LSTM states computed for target word $e_{a_{j-1}}$, effectively encoding the full context around $e_{a_{j-1}}$.

\item Character-based representations on both sides \model{NNJumpBoth}: we consider a more complex alignment model, which in addition takes into account the source side. Using the same representations as for the target side, we make the jump value also depend on the previously aligned source word. The source and target representations are concatenated before being passed through the MLP. 
\begin{equation}
p(a_j| f_{1}^{j-1}, a_1^{j-1}, e_1^I) = p(\Delta_{a_{j}} | [h_{a_{j-1}}, h^{'}_{j-1}])
\end{equation}
where $h^{'}_{j-1}$ is a context-dependent representation of the source word $f_{j-1}$.
\end{itemize}
Again, as source and target words are fully observed, these modifications have no impact on the computations used to compute the various quantities required for the estimation of our models. Finally note that in our implementation, the alignment and the translation models do not share any parameter.
%\begin{figure}[!h]
%\centering
%\begin{tikzpicture}
%\node (E1) at (-60pt, 0pt) {$e_{i-1}$};
%\node (E2) at (-22pt, 0pt) {$e_i$};
%\node (E3) at (0pt, 0pt) {$e_{i+1}$};
%\node (F1) at (22pt, 0pt)  {$f_{i-1}$};
%\node (F2) at (60pt, 0pt)  {$f_i$};
%\node[align=left] (LookupLayer) at (-98pt, -4pt) {\small Character \\ \small hidden state};
%
%\pic [local bounding box=M1] at (-60pt, -12pt) {hidden};
%\pic [local bounding box=M2] at (-30pt, -12pt) {hidden};
%\pic [local bounding box=M3] at (0pt, -12pt) {hidden};
%\pic [local bounding box=M4] at (30pt, -12pt) {hidden};
%\pic [local bounding box=M5] at (60pt, -12pt) {hidden};
%\node[align=left] (HiddenLayer) at (-95pt, -20pt) {\small Bi-LSTM};
%\draw[-latex] (E1)  -- (M1);
%\draw[-latex] (E2) -- (M2);
%\draw[-latex] (E3) -- (M3);
%\draw[-latex] (F1)  -- (M4);
%\draw[-latex] (F2)  -- (M5);
%
%\draw[<->] (M1)  -- (M2);
%\draw[->] (M3)  -- (M4);
%
%\pic [local bounding box=M3] at (0pt, -55pt) {softmax};
%\draw[-latex] (F1) -- (M3);
%\node[align=left] (HiddenLayer) at (-95pt, -62pt) {\small Softmax};
%
%\node (F3) at (0pt, -75pt) {$p(f|e)$};
%
%\end{tikzpicture}
%\caption{Structure of the character-based neural translation model}
%\label{fig:Character_alignment_model}
%\end{figure}

\subsection{Training neural HMMs with EM \label{ssec:algorithm}}
Our training algorithm mostly follows \cite{Tran2016Proceedings}, where expectation-maximization (EM) is combined with back-propagation to train the neural network(s) models. For a number of training epochs, we repeat the following procedure:
 \begin{enumerate}
   \setlength\itemsep{0.5mm}
 \item For each batch:
   \begin{enumerate}
     \setlength\itemsep{1mm}
   \item Compute the posterior probability of each possible alignment link and the auxiliary function of the EM algorithm;
   \item Improve the auxiliary function by performing one gradient update of the neural network parameters.
   \end{enumerate}
 \item After a fixed number of batches, collect and store the entire translation model and jump width distribution for all sentences in the corpus; update the jump distribution. 
 \end{enumerate}

 The initial parameter values are either random (for \model{IBM-1}) or are initialized with the parameter values of the corresponding \model{IBM-1} models (for the \model{HMM} models).

\section{Experiments} \label{sec:experiments}
\subsection{Set-ups and evaluation protocol}

\subsubsection{Implementation}
Our neural translation models are based on a simple architecture composed of a word embedding layer (64 units)\footnote{In our initial experiments with En:Ro, we found that using a larger number of cells (128 or 256) did significantly improve the AER score after 10 iterations. As for the other meta-parameters, we decided to stick with these baseline values: we assume that the relative differences between models observed in our setting would carry-over, albeit with slighlty different values, for larger models.}, feed-forward layers (each comprising 64 units) with activation function htanh \cite{Yang2013Word}, followed by a drop-out layer and a softmax layer. The contextual models use a context window of size $h=1$, based on the experiments reported in \cite{Tamura2014Recurrent}. For the convolutional models, we apply one small filter of size (3,3) to combine context word embeddings. For the character-based models, the bi-LSTM model also contains 64 units in the embedding layers and in the hidden layers.% \fy{Difficult to explain} \fy{Attention aux embeddings qui sont de taille variable !}

In the alignment model, we consider jump values in the interval $[-5, +5]$. In the neural alignment models, the chararacter embeddings are also 64 dimensional; the hidden layer of the MLP contains 80 cells. In all cases, our optimizer is Adam \cite{Kingma2014Adam} with an initial learning rate of 0.001; the batch size is set to 100 sentences. %\fy{Add: neural alignment models}

We use all sentences of length lower than 50 and a 50K word vocabulary for both the source and target languages; in our experiments with character-based models, the source and/or vocabulary is not constrained. However, training still requires to compute a softmax layer, which we approximate when needed by defining ``batch specific'' vocabularies of 5K words containing all the words in the batch plus the remaining most frequent words.

All parameters of the \model{Giza++} and \model{Fastalign} baselines are set to their default values. Note that the baselines use a complete vocabulary for training, which is much larger than the vocabulary size of the neural models, and gives the discrete models a small edge over their neural counterpart. 
% % The out-of-vocabulary word is denoted UNK.
We train all models for 10 EM iterations.

\subsubsection{Datasets}

Our experiments consider several language pairs all having English on one side. For consistency, our training sets are mostly made of sentences from Europarl \cite{Koehn2005Europarl}: this is the case for French, German and Romanian (in the latter case, we also use the SETIMES corpus used in WMT'16 MT evaluation); for Czech we use the parallel data from News Commentary V11 to reproduce another ``small data'' condition.\footnote{Arguably, larger training datasets exist for Czech and Romanian, which we could use to further improve our results.} Testing use standard test sets when applicable: for French and Romanian we use data from the 2003 word alignment challenge \cite{Mihalcea2003Evaluation}; the German test data is also Europarl, while for Czech we use the corpus described in \cite{Marecek16Czech}.

Basic statistics for these corpora are in Table~\ref{tab:train_corpus_size}. En-Fr and En-De training data is much larger than for En-Ro and En-Cz ($\sim$260K and $\sim$190K respectively). As expected, the vocabulary sizes of the German, Romanian and Czech corpora are substantially greater than the corresponding English, which contains a smaller number of inflected variants. These differences of size are thus a factor considered in the evaluation section.

% \fy{Fix: numbers are right aligned}

\begin{table}[h!]
  \begin{center}
    \begin{tabular}{ | c | c | c |  c |}
      \hline
      Corpus & \# sentence & \multicolumn{2}{c|}{vocabulary} \\
             & pairs & English & Foreign \\
      \hline \hline
      En-Fr & $\sim$1.9M  & 122~580  & 126~052\\
      \hline
      En-De & $\sim$1.7M  & 113~037  & 362~517\\
      \hline
      En-Ro  & $\sim$260K &  77~361 & 120~287\\
      \hline
      En-Cz & $\sim$190K & 74~504 & 156~469 \\
      \hline
    \end{tabular}
    \caption{Basic statistics for the training data}
    \label{tab:train_corpus_size}
  \end{center}
\end{table}

\begin{table}[h!]
  \begin{center}
    \begin{tabular}{ | c | c | c | c | c|}
      \hline
      Corpus & \# sent. &  \multicolumn {2}{c|}{\# tokens} & \# non-null links \\
             & & Eng. & For. & \\
      \hline \hline
      En-Fr & 447  & 7~020 & 7~761 & 17~438\\
      \hline
      En-De & 509 & 10~413 & 9~945 & 10~533\\ 	 
      \hline
      En-Ro  & 246 & 5~455 & 5~315 & 5~991\\ 	 
      \hline
      En-Cz & 2~501 & 59~724 & 52~881 & 67~423\\ 	
      \hline
    \end{tabular}
    \caption{Basic statistics for the test data}
    \label{tab:test_corpus_size}
  \end{center}
\end{table}

% \subsubsection{Evaluation metrics}

 We use Alignment Error Rate (AER) \cite{Och2003Minimum} as a measure of performance. AER is based on a comparison of predicted alignment links with a human reference alignments including sure  (S) and possible (P) alignments links, and is defined as an average of the recall and precision taking into account P and S links. Formally, the AER score is defined as:
 \begin{eqnarray}
 AER = 1 - \dfrac{|A \cap S| + |A \cap P|}{|A| + |S|},
 \end{eqnarray}
 where $A$ is the set of predicted alignments. Out of the four datasets, only Romanian/English does not contain Possible links.

\subsection{Results \label{ssec:results}}
Table~\ref{tab:AER_scores} reports the AER scores of our four baselines (\model{IBM-1}, \model{HMM}, \model{IBM-4} implemented in \model{Giza++} and \model{Fastalign}). These are systematically contrasted to our neural network models (\model{IBM-1+NN}, \model{HMM+NN} and their variants). A first general observation is that almost all neural network models outperform their discrete counterpart, with our best HMM models even outperforming \model{IBM-4} for some language pairs. %\fy{Add alignment models numbers}

\begin{table*}[h!]
\centering
\begin{tabular}{| c | c || c | c || c | c || c | c || c | c |}
\hline

\multicolumn{2}{|c||}{Model} 
& En:Fr & Fr:En 
& En:Ge & Ge:En 
& En:Cz & Cz:En 
& En:Ro & Ro:En \\

\hline \hline

\multirow{6}{*}{IBM-1}
& \multirow{1}{*}{Giza++} 
& 40.09 & 33.90  
& 39.02 & 42.65 
& 45.09 & 48.46
& 56.02 & 53.51\\

& \multirow{1}{*}{NN} 
& 27.95 & {\bf 27.20}  
& 37.63 & 39.21 
& 42.28 & {\bf 40.97}
& 46.39 & 44.90 \\

& \multirow{1}{*}{CtxCc} 
& 27.41 & 28.86  
& 36.40 & 36.30 
& 44.63 & 42.32 
& 49.92 & 43.94 \\

& \multirow{1}{*}{CtxCnn} 
& 27.85  & 27.98  
&  37.16 & {\bf 36.02} 
&  45.57 &  41.91
&  {\bf 46.14} &  {\bf 43.92} \\

& \multirow{1}{*}{NNChar} 
& 28.76 & 31.39  
& 36.21 &  40.88 
& 40.84 & 42.35  
& 50.16 & 48.28 \\

& \multirow{1}{*}{NNCharWord} 
& {\bf 27.03} & 28.33  
& {\bf 35.31} &  40.47
& {\bf 40.27} & 46.19 
& 46.53 & 43.93 \\

\hline \hline
\multirow{1}{*}{IBM-2}
& \multirow{1}{*}{Fastalign}
& 15.18 & 16.23 
& 28.97 & 31.28 
& 25.75 & 25.30 
& 33.36 & 32.91 \\

\hline \hline
\multirow{9}{*}{HMM}
& \multirow{1}{*}{Giza++} 
& 11.99 & 11.97 
& 23.91 & 26.33 
& 27.85 &  30.37 
& 33.36  &  36.38\\

& \multirow{1}{*}{NN} 
& 11.83  & 11.14 
& 26.77  &  29.43 
&  23.49 & 24.06 
&  30.69 & 40.12 \\

& \multirow{1}{*}{CtxCc} 
& 10.38  &  11.57 
& 34.35  &  32.30 
& 24.31 & 25.02 
& 33.83 & 34.82 \\

& \multirow{1}{*}{CtxCnn} 
& 11.63  & 12.70  
& 30.63  & 30.34 
&  24.18 &  22.92 
& 30.86 & 34.82 \\

& \multirow{1}{*}{NNCharTgt} 
& 9.17 & 9.55
& 26.03 & 28.10  
& 16.73 & 24.61 
& 27.54 & \textbf{28.01} \\

& \multirow{1}{*}{NNCharWord} 
& 10.45  & 10.27  
& 24.97  & 29.76  
& \textbf{16.04} & \textbf{22.79} 
& \textbf{25.50}  & 29.18 \\

& \multirow{1}{*}{NNCharBoth} 
&  10.90  & 11.17  
& 27.14 	& 29.31   
& 17.38 	& 28.28
& 28.14 & 31.53  \\

% \cline{2-10}

& \multirow{1}{*}{NNJumpTgt} 
&  \textbf{8.40} 	& \textbf{7.70} 
& 23.78 & 25.39 
& 15.93 	& 25.81
& 28.26 & 29.09  \\

& \multirow{1}{*}{NNJumpBoth} 
&  8.47 	& 7.74 
& \textbf{23.69} & \textbf{24.90} 
& 16.38 & 23.87 
& 26.85 & 29.76   \\

\hline \hline
\multirow{1}{*}{IBM-4}
& \multirow{1}{*}{Giza++} 
& 9.99 & 9.63 
& 21.45 & 23.31 
& 20.92  & 26.49 
& 31.04  & 32.29\\
\hline

\end{tabular}

\caption{AER scores. For each language pair and model, we report the AER of each asymmetrical model. The best score for each language direction / model is in boldface.}
\label{tab:AER_scores}
\end{table*}

Most of the improvement is already achieved by the vanilla NN model, which improves over the baseline for all languages, sometimes for a very large margin, eg.\ -8/9 AER for the neural IBM-1 for the pair Ro:En in both directions. The corresponding gains for the basic neural HMM model are not as large, our best improvement being observed for the Cz:En language pair. The improvements are overall lesser for German: on the one hand, the issues with unknown words are not as bad as for Czech, owing to a larger training set; on the other hand all our NN architectures fail to improve the modeling of alignments of German compounds which typically yield many-to-one alignment links that are poorly predicted; word order differences with English are another area where our models do not help much (see Section \ref{ssec:transition_analysis}). 

Regarding contextual variants, a first observation is that the difference between concatenation and convolutions is limited, typically in the order of 1 AER point; the latter approach seems to be on average the best choice. Comparison with the neural \model{IBM-1} baselines reveal that the contextual version is not always better than the default. The largest gains are observed in small data conditions (Ro:En and Cz:En) when English is on the target side: in this case, the context helps to disambiguate alignment links for English words by improving the translation distribution $p(f_j |a_j, e_{a_j-h}^{a_j+h})$. For instance, we found that the context vastly improved the precision (from 0.58 to 0.74) as well as the recall (from 0.5 to 0.54) in the Ro:En data; in the other direction the change is unsignificant. This effect is less clear for the \model{HMM} model, where contextual models are almost always outperformed by character-based variants.

Models using character-based in the target (with or without word information) also yield significant and consistent gains, especially also in small data conditions. Comparing the two conditions, we see that combining word and character information is not always the best approach, as the pure character-based approach is sometimes even better. Our claim is that this approach should be preferred given a sufficient large dataset (as in the Fr:En condition); when this is not the case, word information, which is easier to train, can also prove helpful. With respect to the neural baseline, the gains are maximal when the morphologically rich language is on the target side: in this situation, character-based representations help to differentiate the translation model for the rare words, which in the baseline versions all correspond to the same UNK symbol.\footnote{Remember that the neural models, contrarily to the discrete models, use a limited vocabulary of 50K words.} The use of character models in the target did not enable us to improve these results. % \fy{Why ?}

Regarding alignment models, we see a gain in using a neuralized version of the jump model in the cases where character-based models are already helping, ie.\ for the large data conditions (Fr:En and Ge:En). For the other languages, we do not find any improvements in our setting, probably because the small data condition makes character-based models less effective.
% \fy{Character in target, Alignments}

All in all, using our best models, we obtain symmetrized alignments\footnote{Using the \textsl{grow-diag-final} heuristic proposed in \cite{Koehn05Edinburgh}.} that greatly outperform their corresponding baselines, by 5/6 AER point for Cz:En and Ro:En. Even better scores are obtained when symmetrization uses the best model in each direction: doing so in Ro:En with our best HMM models brings us an additional improvement of about +1 AER (24.93 instead of 25.89). Finally note that all these results were obtained using a limited vocabulary of 50k words for each language; increasing the vocabulary size would be another (computationally expensive) way to further boost alignment quality. % \fy{check this}

\section{Error Analysis \label{sec:error_analysis}}
In this section, we perform a detailed analysis of the quantitative results presented above, focusing mostly on the differences between discrete and neural versions of the \model{HMM} and \model{IBM} model. Our goal in this section is to better understand the improvements brought by the neural models, but also to highlight the problems that remain difficult for alignment models. To this end, we study the error types of each translation model, broken down by link category, where we distinguish links joining frequent vs rare words or unknown words, null links, etc. We also study the difference between the inferred distribution of jumps wrt.\ the actual jump distribution.

\subsection{Issues with unaligned words}
\begin{figure*}[h]
   \centering
   \includegraphics[width=0.9\textwidth]{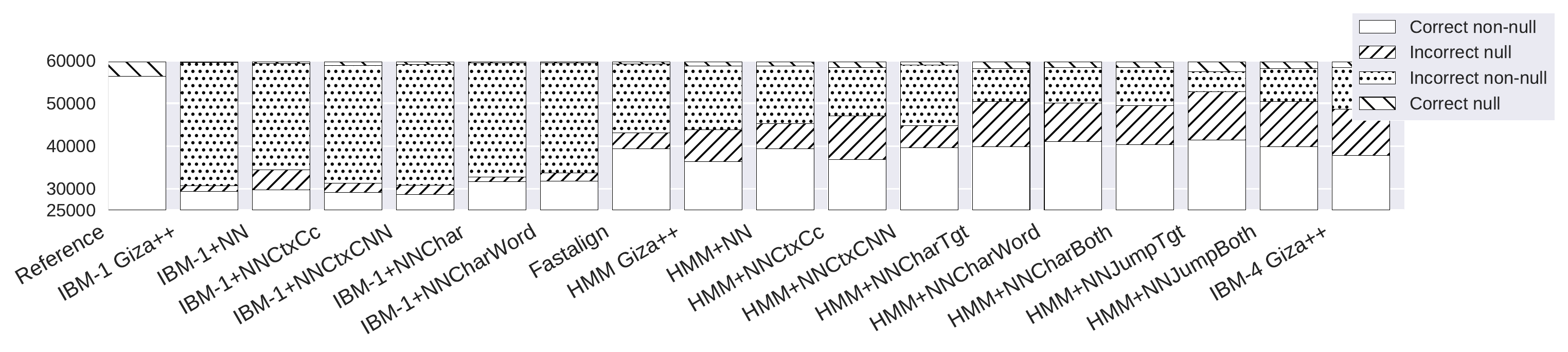}
   \caption{Detailed accurracy of alignment models for En:Cz.}
   \label{fig:accuracy_en_cz}
\end{figure*}

We study the accuracy of alignment models. Figure~\ref{fig:accuracy_en_cz} compares our 17 models for the task of aligning Czech words with their English counterparts: each model makes exactly as many prediction as their are English tokens (see Table~\ref{tab:test_corpus_size}), and these predictions can be broken down into four categories: correct or incorrect null links;\footnote{Cases where a Czech word is aligned with the dummy null English word. In this analysis, null links for English words are not taken into account.} correct or incorrect non-null links). % \fy{What about playing with $p_0$ ?}
As can be observed, models of the \model{IBM-1} family generate very few null links, and concentrate all their efforts in generating correct (or wrong) links between actual words as already noted by \cite{Moore04improving}. The variants of the \model{HMM} model display a different pattern: (a) they make less predictions (and less errors) for non-null links; (b) they tend to predict a large number of null links, with only a small portion of them being actually correct.
About half of the remaining errors of our best models concern null links, in this case the prediction of a link for a word that should have stayed unaligned. Null links are often due to deep to syntactic divergences between languages and are quite hard to predict based on the sole source (or target) word. This is mostly a modeling issue, for which the transition from discrete to neural models is of little help. 
 Similar trends were observed for the other language pairs / directions.
 
\subsection{Issues with rare words and part of speech}
%\fy{Look at Czech:En}
 
\begin{figure}[h]
   \centering
   \includegraphics[width=0.5\textwidth]{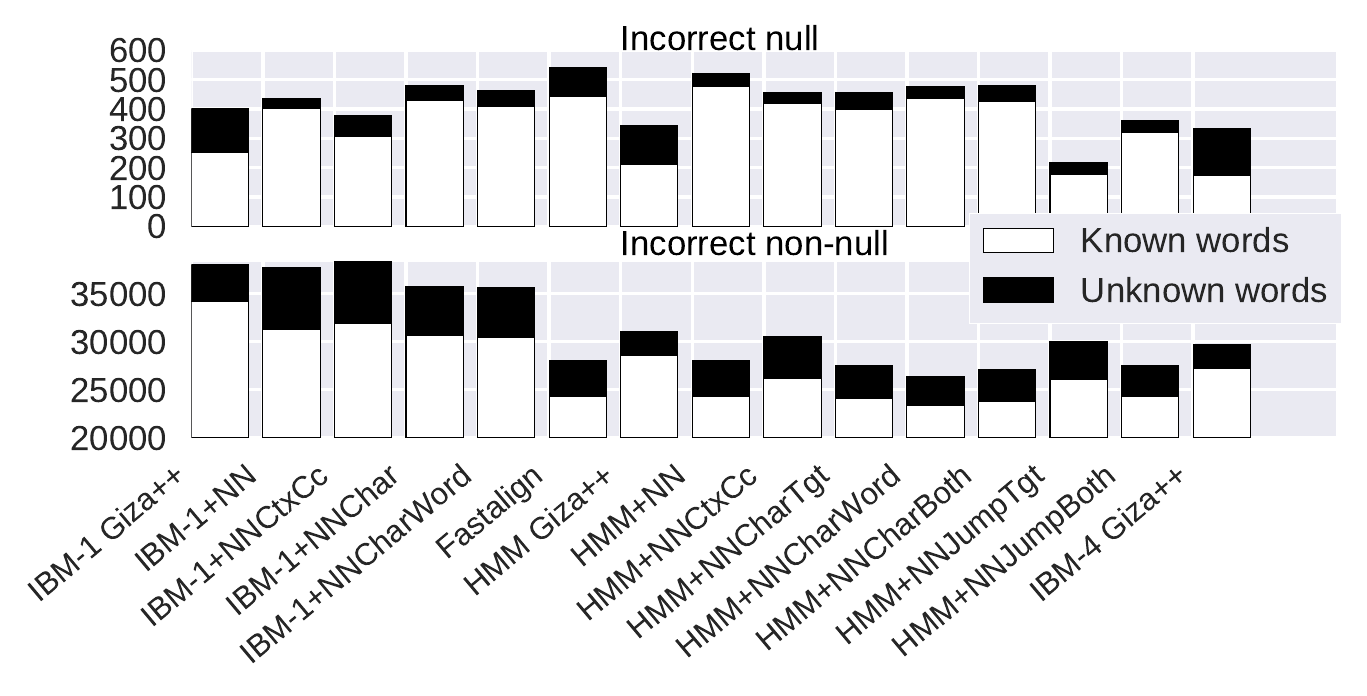}
   \caption{Alignment recall errors for known and unknown words.}
   \label{fig:accuracy_unk_en_cz}
 \end{figure}

A similar view emerges from the analysis of recall: for this, we consider all the En:Cz links that actually exist in the reference, including null-links (or equivalently, non aligned words) and study the alignment patterns that are not present in the model's predictions. We break down the results in two categories: null links and non-null links. The former number correspond to words that should be left unaligned, yet are aligned by the model; the latter correspond to non-null links that are missed in our predictions. In the case of unknown target words (Figure~\ref{fig:accuracy_unk_en_cz}), we see the clear benefits of using neural translation models: both \model{IBM-1} variants and \model{HMM} variants yield a clear reduction of errors, specially null links.
%, specially as concerns unknown words.

\begin{figure}[h]
   \centering
   \includegraphics[width=0.5\textwidth]{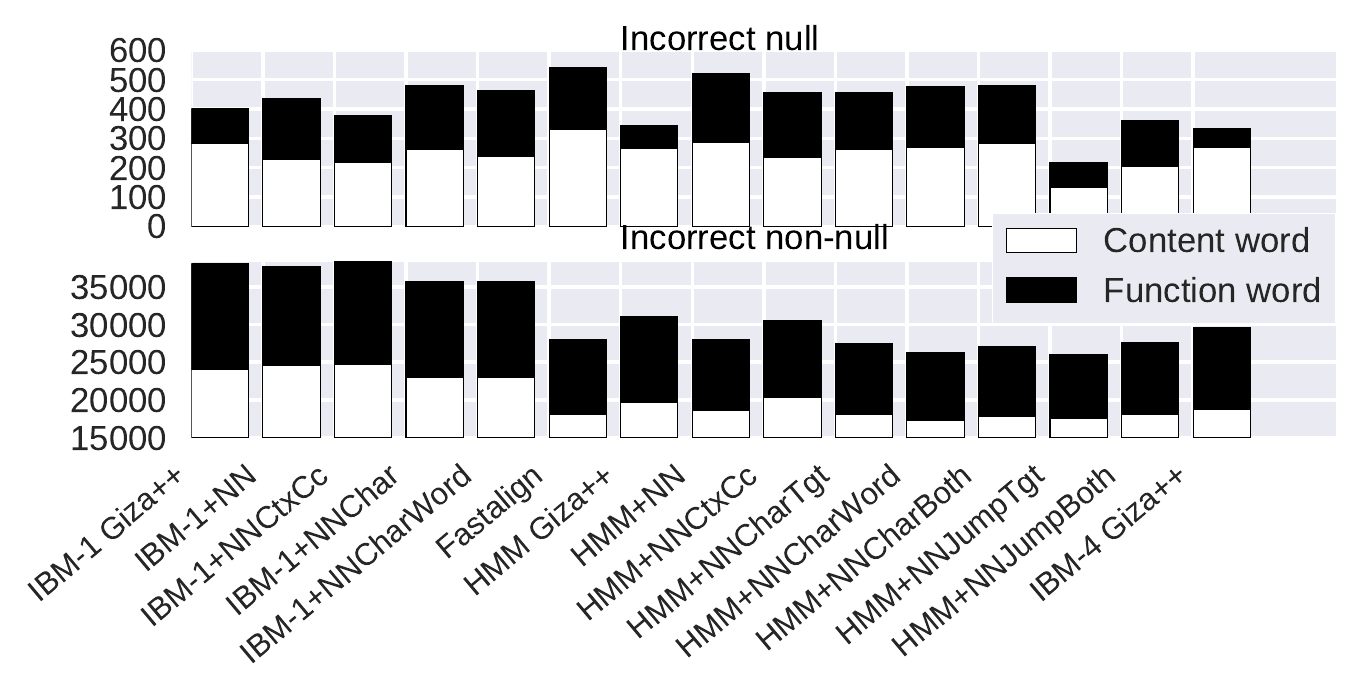}
   \caption{Alignment recall errors broken down by POS}
   \label{fig:accuracy_content_en_cz}
 \end{figure}

The most important gains is obtained with character-based models.
We then categorize target word into two groups of Part-of-Speeches (PoS): content words include noun, verb, adjective and adverb and function words for the remaining PoS (Figure~\ref{fig:accuracy_content_en_cz}). The main observation here is that content words benefit from neural network models whereas the errors for function words is almost unchanged.

% \subsection{Ambiguity with contextual word embeddings}
% We analyze the ambiguity level of contextual translation models \model{NNCtxCc} and \model{NNCtxCnn} over \model{NN} vanilla models by collecting number of distinct source word aligned by each distinct target word. We count the number of source word that \model{NNCtx} models remove from the result of \model{NN}. We show the percentage of correct elimination in Figure~\ref{fig:ambiguity_en_ro}, which demonstrates that the contextual word embedding of English disambiguates more effectively the Ro:En alignment link with the figures being over 68\%.

% \begin{figure}
%   \centering
%   \includegraphics[width=0.5\textwidth]{figs/EnRo_percentage_decrease_context_model}
%   \caption{Percentage of decreasing ambiguity by contextual translation models compared with context-independent models}
%   \label{fig:ambiguity_en_ro}
% \end{figure}

\subsection{Analysis of the transition model \label{ssec:transition_analysis}}
\begin{figure}[h]
  \centering
  \resizebox{0.49\textwidth}{!}{
  \begin{tabular}{cc}
    {\fontsize{15}{15} \selectfont \model{Fastalign}} & {\fontsize{15}{15} \selectfont\model{HMM+NNCharTgt}} \\
    \includegraphics{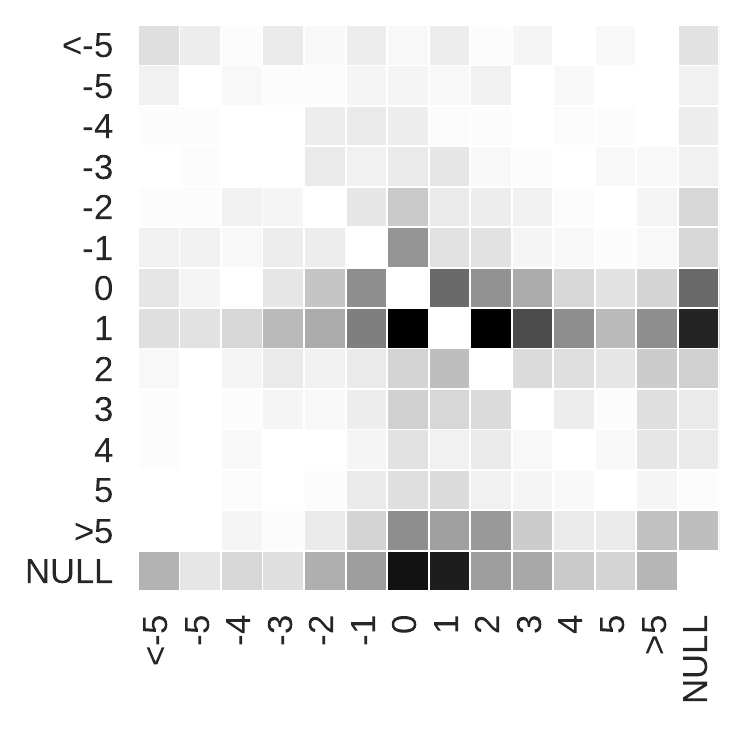} & \includegraphics{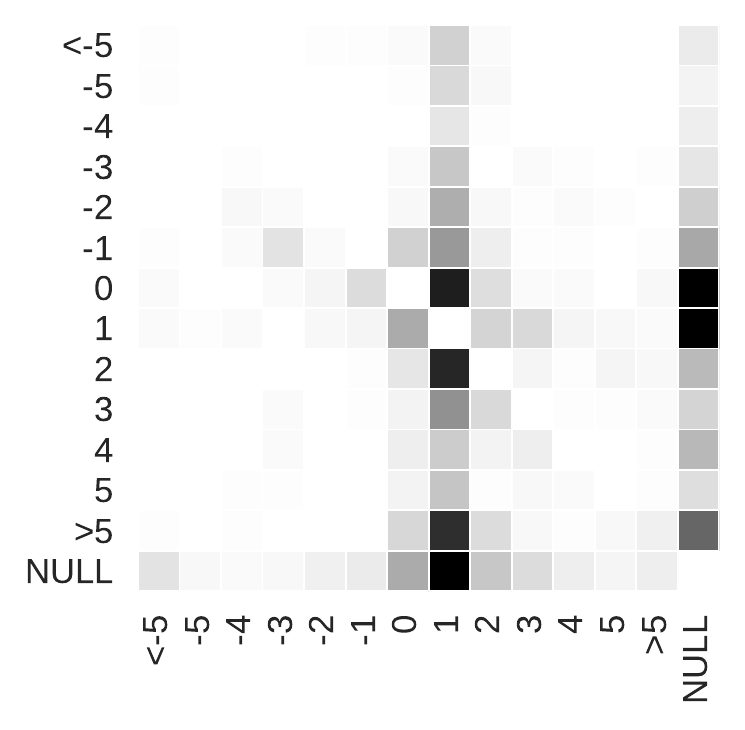} \\ 
    \includegraphics{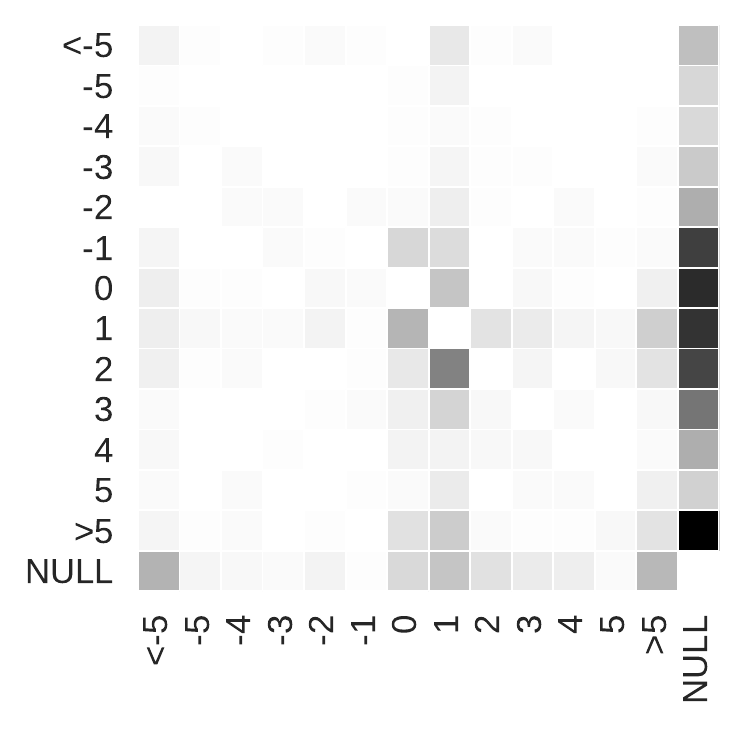} & \includegraphics{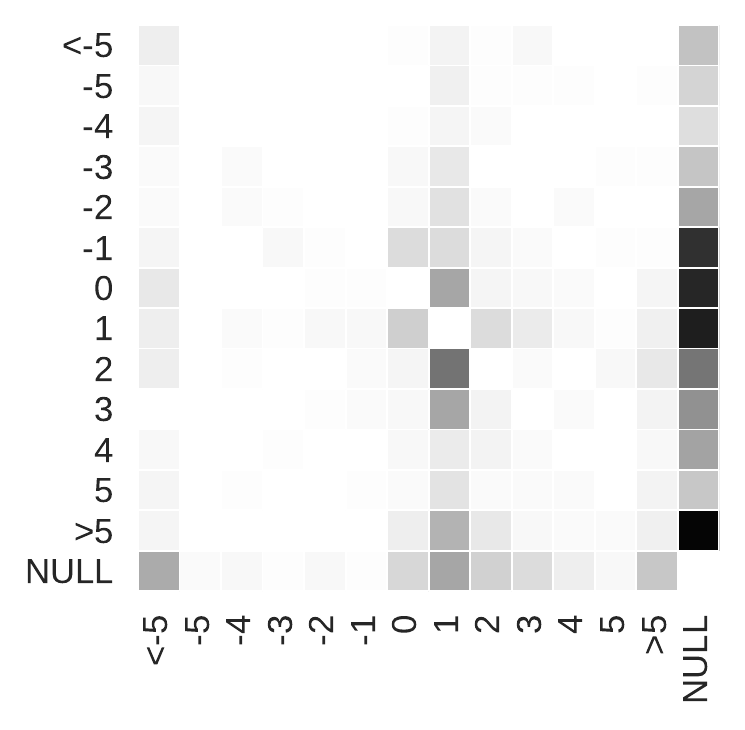} \\
     {\fontsize{15}{15} \selectfont\model{HMM+NNJumpTgt}} & {\fontsize{15}{15} \selectfont\model{HMM+NNJumpBoth}} \\
  \end{tabular} }
  \caption{Confusion matrices for the distortion models. Lines correspond to reference jumps (from $<-5$ to $>+5$ moving upwards), columns to the models predictions (from $<-5$ to $>+5$ moving rightwards).}
  \label{fig:errors_distortions}
\end{figure}

%In our implementations of neural translation models, we keep the distortion model unchanged, which allows us to single out the effect of using a stronger translation model on distortion errors. 
To analyze the distortion errors, we plot the confusions of the distortion models. In these representations, each cell $(k,k')$ counts the number of times the model predicted a jump of $k$ position, whereas the reference jump for that position was $k'$.
\footnote{Determining the ''reference'' jump is a complex issue, as the reference may contain cases of many-to-one alignments, where a target word links to several sources, yielding a set of possible reference jump values. In our analysis, we use the median of all possible target word locations to calculate jump values. We only count an error for each missing or erroneous jump value if the previous target word location is correctly predicted.}
%\footnote{Determining the ``reference'' jump is a complex issue, as the reference may contain cases of many-to-one alignments, where a target word links to several sources, yielding a set of possible reference jump values. In our analysis, we compare the predicted jump with the set of possible values, and count an error for each missing or erroneous jump value. }
These matrices are represented as heatmaps for four alignment models for the language pair En:De, displaying patterns that we also observe for other language pairs (Figure~\ref{fig:errors_distortions}): the darker cell, the greater number of confusions.
%Comparing the two topmost models, we see that \model{IBM-1} over-generates links in three areas: short jumps (with jumps equal to 0 or 1), and long jumps, greater that 5 positions in either directions. Its neuronal counterpart amplifies this tendency to over-predict long jumps. 
On the top part of the graph, we see that in comparison to \model{Fastalign}, the neuronal \model{HMM} models tends to generate a much larger number of short jumps, as well as extraneous null alignments. These problems are somewhat attenuated by our two neural alignment models, which however do not succeed in improving the overall alignment performance.
%This suggests that much remains to be done in terms of better modeling the distortion, our best models having a tendency to concentrate the link distribution around short jumps, a likely sign of a too confident translation model.

%\fy{Add neural alignments, remove IBM1}

\subsection{Garbage collector problem} \label{ssec:garbage_collector_analysis}
One well known issue with \model{Giza++} and \model{Fastalign} is the so-called ''garbage collector problem'' causing rare words in the target language to be misaligned to unfrequent source translations of frequent target words \cite{Brown1993But,Moore04improving,Wang2015Leave}. As a general rule, rare source words should with high probability align to rare targets \cite{Lardilleux11contribution}. To observe this problem, we collect errors corresponding to non-null links between a rare target word and a more frequent source. On the source side, we distinguish into three groups: highly frequent word (accounting overall for 90\% of the training tokens), less frequent words (accounting for 10\% of the tokens),  and unknown words, (never seen in training). On the target side, we define the rare words group as accounting for 1\% of the training data. These errors are reported in table~\ref{tab:errors_freq} for the regular \model{IBM-1} model and the \model{IBM-1+NN}.
As can be observed, the number of errors due to rare target words aligning with more frequent source words is much lower for the neural model, suggesting that it provides a remedy to this problem. This is also illustrated by the example of the French rare word "liquidé" (found only 8 times in the training set), which is misaligned by \model{IBM-1} to common English words such as "had", "see", "taken", "over" and "result". When using \model{IBM-1+NN}, "liquidé" is misaligned only to the English "see".

\begin{table}[htbp]
  \begin{center}
    \begin{tabular}{| c  c | c  c  c  c |}
      \hline
      \multicolumn{2}{|c|}{Model} & \multicolumn{2}{c|}{\model{IBM-1 Giza++}}
      & \multicolumn{2}{c|}{\model{IBM-1+NN}}\\
      \hline
      \multicolumn{2}{|c}{Target} & 1\% & 0\% & 1\% & 0\% \\
      \hline
      \multirow{3}{*}{\rotatebox{90}{Source}}
                                  & 90\% & 131 & 90 &74 & 6 \\
                                  & 10\% & 312 & 167 & 37 & 31 \\
                                  & 0\% & 37 & 46& 0 & 56 \\
      \hline
    \end{tabular}
    \caption{Incorrect non-null links between rare and frequent words (see text for comments).}
    \label{tab:errors_freq}
  \end{center}
\end{table}

\section{Related work \label{sec:related_work}}
With the rapid dissemination of attention-based Neural Translation architectures, which dispenses with the word/phrase alignment step, only a small number of studies have considered this task. 

Early work on neural alignment model is in \cite{Yang2013Word}, which considers a feed-forward network to replace (and generalize) conventional count-based translation model in a HMM model. This line of work is continued by \cite{Tamura2014Recurrent} who show an improvement by using recurrent neural network. Their works aim to improve the alignment quality for a phrase-based translation system by using non-probabilistic scores. \cite{Legrand2016Neural} tackles the problem differently by directly extracting word alignment matrix without using any underlying probabilistic model; this simple symmetrical approach has also proven useful for phrase-par cleaning \cite{Pham18fixing}.  All these works report AER scores and show improvements with respect to standard models, but lack a detailed analysis of the benefits of neural models in alignments.

A much more productive line of research tries to exploit the conceptual similarity between word alignments and attention \cite{Koehn17sixchallenges} with the goal to improve NMT. This can be achieved in several ways: \cite{Cohn16alignment} modify the attention component to integrate some structural bias that have proved useful for alignements, such as a preference for monotonic alignements, for reduced fertilities, etc; they also propose, following \cite{Liang2006Alignment}, to enforce symmetrization constraints, an idea also explored in \cite{Cheng16agreement}; the same general methodology is explored in \cite{Luong15effective,Yang17neural} with the objective to introduce dependencies between adjacent alignment vectors.

The work of \cite{Alkhouli16alignmentbased,Wang17Hybrid} takes a different path, and explore ways to explicitely model alignments in NMT, revisiting with novel tools early word-based translation systems; in their approach, they study various neuralizations, some very similar to our word-based models, of the standard alignment models, and also consider effective training strategies also exploiting weak supervision from count-based models. This line of research is pursued by \cite{Deng18latent}, where attention vectors are (duely) processed as latent variables in NMT. The work of \cite{Rios18deepgenerative} also exploits neural versions of conventional alignment (IBM-1/2) models, with the goal to improve word representations in low resource contexts; contrarily to most work focusing on NMT, some AER scores are reported, which are mostly in line with our baseline neural \model{IBM-1}.

\section{Conclusion and outlook \label{sec:perspective_conclusion}}
In this paper, we have studied alignment models, replacing the traditional count-based translation and alignment models with several variants neural networks, notably contextual models and character-based models. We concentrate on word alignment which provides the base for translation lexicon induction, word sense disambiguation, word noise detection and also machine translation. We observe the performance of our models in word alignment for four language pairs (English versus French, German, Czech and Romanian) and discuss how neural network overcomes alignment difficulties of \model{Giza++} and \model{Fastalign}. One important observation is that neural models can help achieve remarkable improvements in AER scores for most languages pairs, with the higher gains observed for Czech and Romanian, two morphologically rich languages, in a small data condition.
We also show that most of these gains are due to a decrease of non-null link errors. Moreover, our analysis suggests that the alignment problem is still far from solved, and that progress still needs to be made in the prediction of null words on the one hand, and in a more fine grained prediction of jumps on the other hand. We intend to keep working in this direction, trying to close the remaining gap that we observe between well aligned language pairs (say, En:Fr), and pairs that include more distant languages, one of them possibly morphologically complex. One obvious way to progress in this direction is to use better embeddings on the target side or embeddings that are pre-trained on very large monolingual corpora. 
%One obvious way to progress in this direction is to use better embeddings on the target side, using either character-based representations on the target, or embeddings that are pre-trained on very large monolingual corporal.  

Another area where we intend to develop our work is to revisit and improve models that yield symmetrical or near symmetrical alignments \cite{Liang2006Alignment}: in this area, we intend to investigate recent proposals based on variational autoencoders \cite{Kingma14autoencoding}, that have proven effective in various other unsupervised learning tasks.

%----------------------------------------------------------------------------------------
%	BIBLIOGRAPHY
%----------------------------------------------------------------------------------------
\bibliography{Neural_Network_Alignment}

%----------------------------------------------------------------------------------------

\end{document}